\documentclass[conference]{IEEEtran}

\makeatletter
\def\ps@IEEEtitlepagestyle{%
	\def\@oddfoot{\mycopyrightnotice}%
	\def\@evenfoot{}%
}
\def\mycopyrightnotice{%
	{\footnotesize XXX-X-XXXX-XXXX-X/XX/\$XX.00~\copyright~20XX IEEE\hfill}%
	\gdef\mycopyrightnotice{}
}
\makeatother

\usepackage{blindtext}
\usepackage{eso-pic}
\IEEEoverridecommandlockouts
\usepackage{cite}
\usepackage{booktabs,multirow}
\usepackage{graphicx}
\graphicspath{{figures/}}
\usepackage{microtype}
\usepackage{xspace}
\usepackage[hidelinks]{hyperref}
\usepackage{amsmath,amssymb,amsfonts}
\usepackage{algorithm}
\usepackage{algpseudocode}
\usepackage{textcomp}
\usepackage{xcolor}
\def\BibTeX{{\rm B\kern-.05em{\sc i\kern-.025em b}\kern-.08em
		T\kern-.1667em\lower.7ex\hbox{E}\kern-.125emX}}

\usepackage{eso-pic}
\newcommand\AtPageUpperMyright[1]{\AtPageUpperLeft{%
		\put(\LenToUnit{0.17\paperwidth},\LenToUnit{-2cm}){%
			\parbox{0.9\textwidth}{\raggedleft\fontsize{8}{11}\selectfont #1}}%
}}%
\newcommand{\conf}[1]{%
	\AddToShipoutPictureBG*{%
		\AtPageUpperMyright{#1}
	}
}

\newcommand{\argminop}{\mathop{\mathrm{arg\,min}}}
\newcommand{\R}{\mathbb{R}}
\newcommand{\I}{\mathbb{I}}

\begin{document}
	\title{\vspace*{1cm}POSSE-$k$NN: Pathwise Out-of-Bag Selected\\
		Subspace Ensembles for Binary Classification}
	
	\author{\IEEEauthorblockN{Zardad Khan, Amjad Ali, Najd Adeed and Saeed Aldahmani}
		\IEEEauthorblockA{\textit{Department of Analytics in the Digital Era, College of Business and Economics}\\
			\textit{United Arab Emirates University}\\
			\textit{Al Ain 15551, United Arab Emirates}\\
			\{zaar, amjadali, 700047143, saldahmani\}@uaeu.ac.ae}}
	
	\maketitle
	\conf{\textit{Proc. of International Conference on Electrical and Computer Engineering Researches (ICECER 2026) \\
			3-5 December 2026, Istanbul - T\"urkiye}}
	
	\begin{abstract}
		Nearest neighbour classification is attractive for tabular data, but its performance can deteriorate when a fixed query centred neighbourhood does not follow the local class geometry. This study evaluates POSSE-$k$NN, a pathwise $k$ nearest neighbour ensemble that combines bootstrap sampling, random feature subspaces, out-of-bag (OOB) screening, and selective voting. Within each randomized candidate, pathwise selection first chooses the training observation nearest to the query and then chooses each subsequent neighbour relative to the observation accepted at the preceding step. After the candidate is fully specified, its OOB error is computed. Five hundred candidates are generated, ranked by OOB error, and the best 25\% are retained. The method is evaluated on ten binary benchmark datasets using repeated 70/30 train/test partitions and six established comparators. Across the dataset level means, POSSE-$k$NN attains an accuracy of 0.740, Cohen's kappa of 0.412, and a Brier score of 0.175, giving the best aggregate result for all three criteria. It has the highest unrounded mean accuracy and kappa on eight datasets; R$k$NN and SVM lead on the other two. A neighbourhood size analysis on three datasets shows stable behaviour for $k\in\{3,5,7\}$ when the path geometry is informative and identifies a case in which alternative neighbourhood rules are preferable.
	\end{abstract}
	
	
	\section{Introduction}
	The $k$ nearest neighbour ($k$NN) classifier assigns a query to the class represented among its closest training observations. It is nonparametric, transparent, and directly applicable to many tabular learning problems, but its behaviour depends strongly on the distance metric, neighbourhood size, feature representation, and local sample geometry \cite{cover1967nearest,cunningham2021knn}. A conventional neighbourhood is centred permanently at the query. This construction can be restrictive when observations from one class form curved or elongated structures, because later useful observations may be connected through short local steps without belonging to the query's smallest Euclidean ball.
	
	Recent research has addressed these limitations by adapting the neighbourhood, estimating feature relevance, changing the voting rule, or combining several local classifiers. Adaptive evidential $k$NN integrates neighbourhood search and feature weighting \cite{gong2023adaptive}; local mean fuzzy models reduce sensitivity to outliers and class imbalance \cite{kumbure2024local,amer2025localmean}; and feature importance weighted distances alter the contribution of individual coordinates to proximity calculations \cite{addou2026feature}. Double weighted $k$NN jointly weights features and neighbours for high-dimensional genomic classification \cite{ali2025double}. These developments confirm that the definition of a useful local neighbourhood remains an active research problem.
	
	Recent studies further illustrate the breadth of current refinements. Style-linear $k$NN models explicit sample styles while constructing local linear representations \cite{zhang2024style}; proximity-ratio $k$NN reduces the influence of overlapped or noisy observations and compensates for class imbalance \cite{amer2025effective}; and fuzzy neighbourhoods have been strengthened through ensemble feature selection \cite{lohrmann2025benefit} and ordered-weighted aggregation of class-local means \cite{kumbure2025generalizing}. In parallel, optimized model-agnostic random-subspace ensembles learn feature-selection probabilities rather than sampling every coordinate uniformly \cite{huynhthu2024optimizing}. Together, these studies motivate combining adaptive local geometry with controlled feature diversity and internal screening.
	
	A complementary strategy constructs many local classifiers and retains only reliable members. Bagging and random subspaces create diversity through samples and features \cite{breiman1996bagging,ho1998subspace}. Exact bootstrap $k$NN, random $k$NN, and OOB selected $k$NN ensembles further demonstrate that local methods benefit from resampling and internal model screening \cite{steele2009exact,li2014random,gul2018ensemble}. More recent ensembles combine neighbour classifiers through centroid displacement, feature weighting, pathwise neighbourhood retrieval, and alternative representations \cite{wang2023centroid,gul2023selection,ali2023pathwise,ali2024projection}.
	
	POSSE-$k$NN integrates pathwise selection, OOB assessment, random subspaces, selective retention, and ensemble voting. This study makes four contributions. First, it gives a coherent two-stage formulation in which pathwise selection constructs each candidate neighbourhood and OOB error subsequently selects the candidate models. Second, it evaluates the complete method on ten binary datasets against six established classifiers using three complementary criteria. Third, it reports the full repeated holdout accuracy distributions in addition to dataset level means. Fourth, it examines $k\in\{3,5,7\}$ on three datasets to characterize neighbourhood size sensitivity and identify settings in which alternative neighbourhood rules remain competitive.
	
	\section{Proposed Method}
	The proposed procedure has two distinct stages. First, each randomized candidate learner applies the pathwise selection construction to obtain an ordered, query-specific sequence of $k$ neighbours. Second, after the candidate learner has been completely defined, its predictive error is evaluated on its OOB observations and the best candidates are retained for the final ensemble, following the model-selection principle based on OOB performance. Thus, OOB information does not modify a query neighbourhood; it is used only to rank complete pathwise learners.
	
	\subsection{Notation and randomized candidate learners}
	For a positive integer $m$, write $[m]=\{1,\ldots,m\}$. Let the preprocessed training data be
	\begin{equation}
		\mathcal D=\{(\mathbf x_i,y_i):i\in[n]\},\qquad \mathbf x_i\in\R^p,\quad y_i\in\{0,1\}.
		\label{eq:data}
	\end{equation}
	The preprocessing map is estimated from the current training partition and then applied unchanged to OOB and test observations. Let $B$ be the number of candidate learners, $k\leq n$ an odd neighbourhood size, $p'\in[p]$ the number of features per candidate, $q\geq1$ the order of the Minkowski distance, and $\rho\in(0,1]$ the proportion of candidates retained after OOB evaluation.
	
	For candidate $b\in[B]$, draw the bootstrap index sequence
	\begin{equation}
		\mathbf J_b=(J_{b1},\ldots,J_{bn}),\qquad J_{br}\overset{\mathrm{iid}}{\sim}\operatorname{Unif}([n]),
		\label{eq:bootstrap}
	\end{equation}
	where $r\in[n]$ denotes a position in the bootstrap sample. The corresponding ordered bootstrap sample is $\mathcal D_b^*=\bigl((\mathbf x_{J_{br}},y_{J_{br}})\bigr)_{r=1}^{n}$; this sequence notation preserves repeated bootstrap rows. Let
	\begin{equation}
		S_b=\{J_{br}:r\in[n]\},\qquad O_b=[n]\setminus S_b
		\label{eq:oobsets}
	\end{equation}
	be, respectively, the set of distinct in-bag indices and its OOB complement. If $O_b=\varnothing$, the bootstrap sequence is redrawn. Independently, select a feature set $F_b\subseteq[p]$ uniformly without replacement, with $|F_b|=p'$. Candidate $b$ is therefore determined by the fixed pair $(\mathbf J_b,F_b)$.
	
	Distances for candidate $b$ are computed only on $F_b$:
	\begin{equation}
		d_{b,q}(\mathbf u,\mathbf v)=\left(\sum_{j\in F_b}|u_j-v_j|^q\right)^{1/q},\qquad \mathbf u,\mathbf v\in\R^p.
		\label{eq:distance}
	\end{equation}
	The experiments use $q=2$, so Eq.~\eqref{eq:distance} reduces to Euclidean distance in the selected subspace. Reserving $q$ for the norm order avoids confusing the distance parameter with a bootstrap position or a path index.
	
	\subsection{Stage 1: Pathwise neighbourhood selection}
	For a query $\mathbf z\in\R^p$, initialize the moving reference point and the available bootstrap positions as
	\begin{equation}
		\mathbf v_{b,0}(\mathbf z)=\mathbf z,\qquad C_{b,0}(\mathbf z)=[n].
		\label{eq:init}
	\end{equation}
	For steps $t=1,\ldots,k$, select one bootstrap position and update the reference point recursively:
	\begin{align}
		r_{b,t}(\mathbf z)&\in\argminop_{r\in C_{b,t-1}(\mathbf z)}d_{b,q}\!\left(\mathbf v_{b,t-1}(\mathbf z),\mathbf x_{J_{br}}\right), \label{eq:pathselect}\\
		\mathbf v_{b,t}(\mathbf z)&=\mathbf x_{J_{b,r_{b,t}(\mathbf z)}}, \label{eq:pathupdate}\\
		C_{b,t}(\mathbf z)&=C_{b,t-1}(\mathbf z)\setminus\{r_{b,t}(\mathbf z)\}. \label{eq:pathavailable}
	\end{align}
	Equation~\eqref{eq:pathselect} defines the pathwise sequence: the first neighbour is closest to $\mathbf z$, whereas neighbour $t\geq2$ is closest to the neighbour chosen at step $t-1$. Equation~\eqref{eq:pathavailable} removes the selected bootstrap position so that the rule returns $k$ positions rather than repeatedly selecting the same row. Because bootstrap sampling is with replacement, two distinct positions can still contain the same original observation; such duplicates correctly retain their bootstrap multiplicity. Distance ties are resolved by a fixed deterministic convention, such as choosing the smallest eligible bootstrap position.
	
	The ordered pathwise neighbourhood and the candidate vote are
	\begin{align}
		P_b(\mathbf z)&=\left(J_{b,r_{b,1}(\mathbf z)},\ldots,J_{b,r_{b,k}(\mathbf z)}\right), \label{eq:pathset}\\
		s_b(\mathbf z)&=\frac{1}{k}\sum_{t=1}^{k}y_{J_{b,r_{b,t}(\mathbf z)}},\qquad h_b(\mathbf z)=\I\{s_b(\mathbf z)\geq1/2\}. \label{eq:candidatevote}
	\end{align}
	Here $s_b(\mathbf z)$ is the class-one proportion along the selected path and $h_b(\mathbf z)\in\{0,1\}$ is the complete prediction of candidate $b$. Since $k$ is odd, the within-candidate majority vote cannot tie. Importantly, Eqs.~\eqref{eq:pathselect}--\eqref{eq:candidatevote} define the neighbourhood and prediction without using OOB performance.
	
	\subsection{Stage 2: OOB model selection and ensemble prediction}
	The same fixed pair $(\mathbf J_b,F_b)$ is used to predict every observation in $O_b$. For each $i\in O_b$, the pathwise-selection recursion is restarted at $\mathbf v_{b,0}(\mathbf x_i)=\mathbf x_i$ and produces $h_b(\mathbf x_i)$. Because $i\notin S_b$, the OOB observation is absent from the bootstrap sample used by candidate $b$. Its OOB misclassification error is
	\begin{equation}
		e_b=\frac{1}{|O_b|}\sum_{i\in O_b}\I\{h_b(\mathbf x_i)\neq y_i\}.
		\label{eq:ooberror}
	\end{equation}
	Let $\sigma$ be a permutation of $[B]$ satisfying $e_{\sigma(1)}\leq\cdots\leq e_{\sigma(B)}$; tied errors are ordered by candidate index. Define
	\begin{equation}
		B'=\max\{1,\lceil\rho B\rceil\},\qquad \mathcal A=\{\sigma(1),\ldots,\sigma(B')\},
		\label{eq:retain}
	\end{equation}
	where $B'$ is the number of selected candidates and $\mathcal A$ is their index set. For a new query $\mathbf z$, only the selected candidates are evaluated. The class-one vote fraction and final prediction are
	\begin{equation}
		\widehat\pi(\mathbf z)=\frac{1}{B'}\sum_{\ell=1}^{B'}h_{\sigma(\ell)}(\mathbf z),\qquad \widehat y(\mathbf z)=\I\{\widehat\pi(\mathbf z)\geq1/2\}.
		\label{eq:vote}
	\end{equation}
	An exact ensemble tie is assigned to class 1 by the displayed weak inequality; any alternative fixed convention must be declared before evaluation. The separation between the two stages is essential: pathwise selection determines which observations form each local neighbourhood, whereas OOB error determines which complete randomized learners are allowed to vote.
	
	\begin{algorithm}[!t]
		\caption{POSSE-$k$NN: pathwise selection followed by OOB model selection}
		\label{alg:posse}
		\footnotesize
		\begin{algorithmic}[1]
			\Require $\mathcal D,B,k,p',q,\rho$ and query $\mathbf z$
			\Function{\textsc{PathwisePredict}}{$\mathbf u,\mathbf J_b,F_b,k,q$}
			\State $C\gets[n]$, $\mathbf v\gets\mathbf u$, $c\gets0$
			\For{$t=1,\ldots,k$}
			\State $r_t\gets\argminop_{r\in C}d_{b,q}(\mathbf v,\mathbf x_{J_{br}})$
			\State $c\gets c+y_{J_{br_t}}$; $\mathbf v\gets\mathbf x_{J_{br_t}}$; $C\gets C\setminus\{r_t\}$
			\EndFor
			\State \Return $\I\{c\geq(k+1)/2\}$
			\EndFunction
			\For{$b=1,\ldots,B$}
			\State Draw $\mathbf J_b$ and $F_b$; form $O_b=[n]\setminus\{J_{br}:r\in[n]\}$
			\State $e_b\gets|O_b|^{-1}\sum_{i\in O_b}\I\{\textsc{PathwisePredict}(\mathbf x_i,\mathbf J_b,F_b,k,q)\neq y_i\}$
			\EndFor
			\State Rank $e_b$ increasingly and set $B'\gets\max\{1,\lceil\rho B\rceil\}$
			\State Retain $\mathcal A\gets\{\sigma(1),\ldots,\sigma(B')\}$
			\State $\widehat\pi(\mathbf z)\gets {B'}^{-1}\sum_{b\in\mathcal A}\textsc{PathwisePredict}(\mathbf z,\mathbf J_b,F_b,k,q)$
			\State \Return $\widehat\pi(\mathbf z)$ and $\widehat y(\mathbf z)=\I\{\widehat\pi(\mathbf z)\geq1/2\}$
		\end{algorithmic}
	\end{algorithm}
	
	\begin{table}[!t]
		\caption{Characteristics and data sources of the benchmark datasets.}
		\label{tab:datasets}
		\centering
		\scriptsize
		\renewcommand{\arraystretch}{1.08}
		\setlength{\tabcolsep}{5.0pt}
		\begin{tabular}{clrrcc}
			\toprule
			ID & Dataset & $p$ & $n$ & Class counts & Source \\
			\midrule
			$D_{1}$ & ILPD & 10 & 583 & (167, 416) & \href{https://www.openml.org/d/1480}{OpenML} \\
			$D_{2}$ & Heart & 13 & 303 & (138, 165) & \href{https://archive.ics.uci.edu/dataset/45/heart+disease}{UCI} \\
			$D_{3}$ & Echo Months & 9 & 130 & (64, 66) & \href{https://www.openml.org/d/944}{OpenML} \\
			$D_{4}$ & AR5 & 29 & 36 & (8, 28) & \href{https://www.openml.org/d/1062}{OpenML} \\
			$D_{5}$ & Cleveland & 13 & 303 & (138, 165) & \href{https://www.openml.org/d/40710}{OpenML} \\
			$D_{6}$ & Breast Tumor & 9 & 286 & (120, 166) & \href{https://www.openml.org/d/844}{OpenML} \\
			$D_{7}$ & Wisconsin & 32 & 194 & (90, 104) & \href{https://www.openml.org/d/753}{OpenML} \\
			$D_{8}$ & Tuning SVM & 80 & 156 & (54, 102) & \href{https://www.openml.org/d/41976}{OpenML} \\
			$D_{9}$ & Grub Damage & 8 & 155 & (49, 106) & \href{https://www.openml.org/d/1026}{OpenML} \\
			$D_{10}$ & Chscase Vine & 8 & 52 & (24, 28) & \href{https://www.openml.org/d/815}{OpenML} \\
			\bottomrule
		\end{tabular}
	\end{table}
	
	\section{Experimental Protocol}
	\subsection{Datasets, configuration, and comparators}
	Table~\ref{tab:datasets} summarizes the ten binary datasets and links to their public repositories. The class counts are ordered as (minority, majority). The benchmarks range from 36 to 583 observations and from 8 to 80 predictors, covering small sample, imbalanced, and moderate dimensional settings. Nonnumeric predictors were integer encoded, and all predictors were min--max scaled to $[0,1]$. Performance was recorded over repeated random 70/30 train/test partitions. Nine datasets have 100 complete repetitions, whereas $D_{4}$ has 94.
	
	The POSSE-$k$NN configuration used $B=500$ candidate learners, $p'=\operatorname{round}(\sqrt p)$ predictors per learner, $k=3$, Minkowski order $q=2$ (Euclidean distance), and $\rho=0.25$. Thus, $B'=125$ OOB-ranked pathwise learners entered the final vote. The comparison methods were standard $k$NN, weighted $k$NN (W$k$NN), random $k$NN (R$k$NN) \cite{li2014random}, random forest (RF), optimal trees ensemble (OTE), and a linear support vector machine (SVM).
	
	\subsection{Evaluation criteria}
	Accuracy is the proportion of correct predictions. Cohen's kappa adjusts observed agreement for agreement expected from the class marginals \cite{cohen1960kappa}. Probability error is measured by the Brier score (BS) \cite{brier1950verification},
	\begin{equation}
		\mathrm{BS}=\frac{1}{N_{\mathrm{te}}}\sum_{i=1}^{N_{\mathrm{te}}}\left[\widehat\pi(\mathbf x_i^{\mathrm{te}})-y_i^{\mathrm{te}}\right]^2,
		\label{eq:brier}
	\end{equation}
	where $(\mathbf x_i^{\mathrm{te}},y_i^{\mathrm{te}})$ is the $i$th test pair and lower values indicate better probability estimates. Dataset level means give every benchmark equal weight in the aggregate row. In Tables~\ref{tab:main_results} and \ref{tab:k_sensitivity}, boldface identifies the best unrounded value; higher Accuracy and kappa and lower BS are preferred.
	
	\begin{table*}[!t]
		\caption{Mean predictive performance of the seven classifiers on the ten benchmark datasets.}
		\label{tab:main_results}
		\centering
		\scriptsize
		\renewcommand{\arraystretch}{0.94}
		\setlength{\tabcolsep}{3.05pt}
		\begin{tabular*}{\textwidth}{@{\extracolsep{\fill}}lccccccc@{}}
			\toprule
			\multicolumn{8}{l}{\textbf{Panel A: Accuracy} (higher is better)} \\
			Dataset & \textbf{POSSE-$k$NN} & $k$NN & W$k$NN & R$k$NN & RF & OTE & SVM \\
			\midrule
			$D_{1}$ & 0.717 & 0.678 & 0.682 & \textbf{0.717} & 0.707 & 0.703 & 0.710 \\
			$D_{2}$ & \textbf{0.835} & 0.798 & 0.750 & 0.828 & 0.825 & 0.808 & 0.828 \\
			$D_{3}$ & \textbf{0.746} & 0.677 & 0.707 & 0.732 & 0.711 & 0.693 & 0.698 \\
			$D_{4}$ & \textbf{0.838} & 0.821 & 0.789 & 0.826 & 0.824 & 0.790 & 0.782 \\
			$D_{5}$ & \textbf{0.827} & 0.776 & 0.745 & 0.823 & 0.814 & 0.801 & 0.793 \\
			$D_{6}$ & \textbf{0.581} & 0.525 & 0.530 & 0.577 & 0.571 & 0.544 & 0.577 \\
			$D_{7}$ & \textbf{0.581} & 0.556 & 0.538 & 0.568 & 0.570 & 0.565 & 0.551 \\
			$D_{8}$ & \textbf{0.709} & 0.651 & 0.622 & 0.697 & 0.694 & 0.681 & 0.643 \\
			$D_{9}$ & \textbf{0.793} & 0.772 & 0.720 & 0.751 & 0.769 & 0.749 & 0.772 \\
			$D_{10}$ & 0.775 & 0.782 & 0.756 & 0.750 & 0.779 & 0.763 & \textbf{0.786} \\
			\midrule
			Mean & \textbf{0.740} & 0.703 & 0.684 & 0.727 & 0.726 & 0.710 & 0.714 \\
			\midrule
			\addlinespace[0.55mm]
			\multicolumn{8}{l}{\textbf{Panel B: Cohen's kappa} (higher is better)} \\
			Dataset & \textbf{POSSE-$k$NN} & $k$NN & W$k$NN & R$k$NN & RF & OTE & SVM \\
			\midrule
			$D_{1}$ & 0.111 & 0.186 & \textbf{0.238} & 0.100 & 0.195 & 0.207 & 0.018 \\
			$D_{2}$ & \textbf{0.665} & 0.591 & 0.494 & 0.651 & 0.645 & 0.610 & 0.650 \\
			$D_{3}$ & \textbf{0.490} & 0.355 & 0.412 & 0.466 & 0.422 & 0.386 & 0.397 \\
			$D_{4}$ & \textbf{0.558} & 0.516 & 0.437 & 0.503 & 0.493 & 0.394 & 0.415 \\
			$D_{5}$ & \textbf{0.647} & 0.550 & 0.483 & 0.638 & 0.623 & 0.597 & 0.580 \\
			$D_{6}$ & \textbf{0.139} & 0.027 & 0.035 & 0.120 & 0.128 & 0.075 & 0.137 \\
			$D_{7}$ & \textbf{0.158} & 0.113 & 0.076 & 0.132 & 0.136 & 0.124 & 0.101 \\
			$D_{8}$ & \textbf{0.300} & 0.248 & 0.207 & 0.284 & 0.289 & 0.261 & 0.212 \\
			$D_{9}$ & \textbf{0.502} & 0.471 & 0.374 & 0.349 & 0.447 & 0.410 & 0.447 \\
			$D_{10}$ & 0.547 & 0.559 & 0.504 & 0.500 & 0.552 & 0.518 & \textbf{0.564} \\
			\midrule
			Mean & \textbf{0.412} & 0.362 & 0.326 & 0.374 & 0.393 & 0.358 & 0.352 \\
			\midrule
			\addlinespace[0.55mm]
			\multicolumn{8}{l}{\textbf{Panel C: Brier score (BS)} (lower is better)} \\
			Dataset & \textbf{POSSE-$k$NN} & $k$NN & W$k$NN & R$k$NN & RF & OTE & SVM \\
			\midrule
			$D_{1}$ & 0.177 & 0.218 & 0.318 & \textbf{0.176} & 0.178 & 0.183 & 0.200 \\
			$D_{2}$ & \textbf{0.125} & 0.167 & 0.250 & 0.147 & 0.127 & 0.133 & 0.128 \\
			$D_{3}$ & \textbf{0.178} & 0.227 & 0.293 & 0.179 & 0.186 & 0.206 & 0.221 \\
			$D_{4}$ & \textbf{0.119} & 0.132 & 0.211 & 0.122 & 0.124 & 0.185 & 0.153 \\
			$D_{5}$ & \textbf{0.127} & 0.172 & 0.255 & 0.148 & 0.132 & 0.138 & 0.148 \\
			$D_{6}$ & 0.255 & 0.317 & 0.470 & 0.254 & 0.274 & 0.311 & \textbf{0.239} \\
			$D_{7}$ & 0.254 & 0.308 & 0.462 & 0.252 & 0.254 & 0.262 & \textbf{0.251} \\
			$D_{8}$ & 0.198 & 0.244 & 0.378 & 0.198 & \textbf{0.190} & 0.198 & 0.216 \\
			$D_{9}$ & \textbf{0.155} & 0.186 & 0.280 & 0.164 & 0.166 & 0.181 & 0.165 \\
			$D_{10}$ & 0.164 & \textbf{0.151} & 0.244 & 0.172 & 0.157 & 0.178 & 0.165 \\
			\midrule
			Mean & \textbf{0.175} & 0.212 & 0.316 & 0.181 & 0.179 & 0.198 & 0.189 \\
			\bottomrule
		\end{tabular*}
	\end{table*}
	
	\section{Results}
	\subsection{Main comparison}
	Panel A of Table~\ref{tab:main_results} reports accuracy. POSSE-$k$NN has the highest unrounded mean on $D_2$--$D_9$, giving eight wins in ten datasets. On $D_1$, R$k$NN is higher by only $5.7\times10^{-5}$, so both values round to 0.717; the boldface follows the unrounded values. On $D_{10}$, SVM obtains 0.786 compared with 0.775 for POSSE-$k$NN. Across the ten dataset means, POSSE-$k$NN leads with 0.740, followed by R$k$NN at 0.727 and RF at 0.726.
	
	Panel B shows the same eight versus two pattern for Cohen's kappa. POSSE-$k$NN is best on $D_2$--$D_9$, W$k$NN leads on $D_1$, and SVM leads on $D_{10}$. The aggregate kappa is 0.412 for POSSE-$k$NN and 0.393 for RF. Panel C gives a less uniform probability ranking result. POSSE-$k$NN has the lowest BS on $D_2$--$D_5$ and $D_9$; R$k$NN, SVM, RF, and $k$NN lead on the remaining datasets. Nevertheless, its aggregate BS of 0.175 is lower than RF (0.179) and R$k$NN (0.181).
	
\begin{figure*}[!t]
		\centering
		\vspace{-0.28in}
		\includegraphics[width=0.71\textwidth]{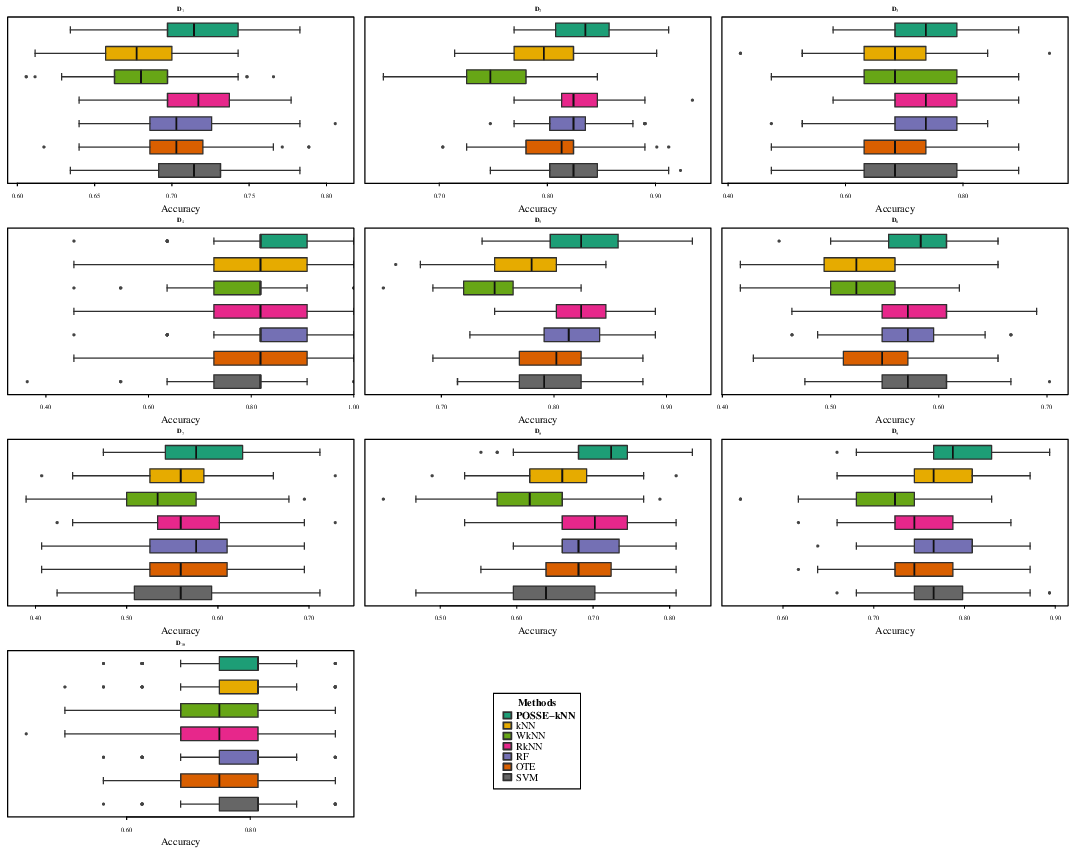}
		\caption{Repeated holdout accuracy distributions for the classifiers on the ten benchmark datasets.}
		\label{fig:accuracy_boxplots}
	\end{figure*}

	Figure~\ref{fig:accuracy_boxplots} shows the complete repeated split accuracy distributions, with dataset identifiers following Table~\ref{tab:datasets}. Each horizontal Tukey box represents the interquartile range, the centre line is the median, whiskers extend to 1.5 times the interquartile range, and circles denote outliers. The POSSE-$k$NN boxes are generally displaced to the right on $D_2$--$D_9$, although overlap with R$k$NN or RF remains substantial on several datasets. The two exceptions are informative. On $D_1$, the POSSE-$k$NN and R$k$NN distributions are almost coincident, but W$k$NN yields a higher kappa, indicating that similar overall accuracy can conceal different class-wise behaviour. On $D_{10}$, SVM has the strongest mean and a compact upper distribution, showing that a linear boundary is preferable for that dataset. These cases prevent the aggregate lead from being interpreted as universal dominance.
	
	\subsection{Sensitivity to neighbourhood size}
	Table~\ref{tab:k_sensitivity} examines neighbourhood size sensitivity on $D_1$--$D_3$. All cells are based on 100 complete repetitions. On $D_2$ (Heart), POSSE-$k$NN is best in accuracy, kappa, and BS for all three values of $k$, and the differences across $k$ are small. On $D_3$ (Echo Months), increasing $k$ from 3 to 7 raises the POSSE-$k$NN accuracy from 0.746 to 0.767 and kappa from 0.490 to 0.534, although R$k$NN gives the lower BS at $k=5$ and 7. The behaviour on $D_1$ (ILPD) is different: accuracy remains near 0.717, but W$k$NN has the highest kappa and R$k$NN the lowest BS. Thus, increasing the path length does not compensate for an unsuitable local geometry in every problem.
	
	\begin{table*}[!t]
		\caption{Sensitivity of the four nearest neighbour classifiers to $k\in\{3,5,7\}$ on $D_{1}$--$D_{3}$.}
		\label{tab:k_sensitivity}
		\centering
		\scriptsize
		\renewcommand{\arraystretch}{0.94}
		\setlength{\tabcolsep}{3.10pt}
		\begin{tabular*}{\textwidth}{@{\extracolsep{\fill}}llccccccccc@{}}
			\toprule
			\multirow{2}{*}{Metric} & \multirow{2}{*}{Method} & \multicolumn{3}{c}{$D_{1}$} & \multicolumn{3}{c}{$D_{2}$} & \multicolumn{3}{c}{$D_{3}$} \\
			\cmidrule(lr){3-5}\cmidrule(lr){6-8}\cmidrule(lr){9-11}
			& & $k=3$ & $k=5$ & $k=7$ & $k=3$ & $k=5$ & $k=7$ & $k=3$ & $k=5$ & $k=7$ \\
			\midrule
			\multirow{4}{*}{Accuracy} & \textbf{POSSE-$k$NN} & 0.717 & 0.716 & 0.717 & \textbf{0.835} & \textbf{0.833} & \textbf{0.838} & \textbf{0.746} & \textbf{0.761} & \textbf{0.767} \\
			& $k$NN & 0.678 & 0.689 & 0.699 & 0.798 & 0.801 & 0.784 & 0.677 & 0.693 & 0.702 \\
			& W$k$NN & 0.682 & 0.683 & 0.683 & 0.750 & 0.786 & 0.808 & 0.707 & 0.702 & 0.700 \\
			& R$k$NN & \textbf{0.717} & \textbf{0.717} & \textbf{0.717} & 0.828 & 0.831 & 0.826 & 0.731 & 0.749 & 0.756 \\
			\midrule
			\multirow{4}{*}{Cohen's kappa} & \textbf{POSSE-$k$NN} & 0.111 & 0.078 & 0.060 & \textbf{0.665} & \textbf{0.661} & \textbf{0.671} & \textbf{0.490} & \textbf{0.519} & \textbf{0.534} \\
			& $k$NN & 0.186 & 0.184 & 0.194 & 0.591 & 0.598 & 0.564 & 0.355 & 0.387 & 0.409 \\
			& W$k$NN & \textbf{0.238} & \textbf{0.221} & \textbf{0.215} & 0.494 & 0.567 & 0.610 & 0.412 & 0.404 & 0.401 \\
			& R$k$NN & 0.100 & 0.076 & 0.063 & 0.651 & 0.658 & 0.649 & 0.463 & 0.503 & 0.519 \\
			\midrule
			\multirow{4}{*}{BS} & \textbf{POSSE-$k$NN} & 0.177 & 0.180 & 0.184 & \textbf{0.125} & \textbf{0.124} & \textbf{0.124} & \textbf{0.178} & 0.180 & 0.185 \\
			& $k$NN & 0.218 & 0.203 & 0.193 & 0.167 & 0.152 & 0.149 & 0.227 & 0.207 & 0.193 \\
			& W$k$NN & 0.318 & 0.258 & 0.250 & 0.250 & 0.158 & 0.140 & 0.293 & 0.277 & 0.270 \\
			& R$k$NN & \textbf{0.176} & \textbf{0.176} & \textbf{0.177} & 0.147 & 0.146 & 0.148 & 0.180 & \textbf{0.178} & \textbf{0.179} \\
			\bottomrule
		\end{tabular*}
	\end{table*}
	
	\section{Discussion}
	The results support the two-stage combination of pathwise selection and OOB model selection, together with random subspaces, while also defining its limits. Pathwise selection first follows a locally connected sequence that a fixed query-centred neighbourhood may miss. Only after each candidate has been defined does OOB ranking suppress candidates that fail on observations excluded from their bootstrap samples. Subspace sampling changes the distance ordering and supplies diverse path hypotheses. The aggregate gains and the eight accuracy/kappa wins are consistent with these mechanisms.
	
	The two non-leading datasets clarify when the design may offer little advantage. On $D_1$, the method matches the best accuracy only after rounding but produces substantially lower kappa than W$k$NN, suggesting sensitivity to class prevalence and the composition of the retrieved path. On Chscase Vine, SVM leads in both accuracy and kappa, while standard $k$NN has the lowest BS. A low-dimensional problem with a comparatively regular boundary may not require a randomized path ensemble, and model screening cannot correct a systematically unsuitable base geometry.
	
	The $k$ sensitivity results also show that neighbourhood size should not be discussed independently of the retrieval rule. Heart is stable across $k=3,5,7$, and Echo Months benefits from a longer path in hard classification measures. ILPD, however, does not improve. In deployment, $k$, the subspace size $p'$, and the retained fraction $\rho$ should therefore be tuned inside the training data. The same principle applies to preprocessing: scaling, imputation, and categorical encoding must be estimated from the training partition and applied unchanged to test observations.
	
	With exhaustive search, predicting one query with one pathwise learner requires at most $O(knp')$ distance work, and the retained ensemble requires $O(B'knp')$. OOB scoring is more expensive but candidate construction is embarrassingly parallel. Retaining 125 of 500 candidates reduces the final voting cost by 75\% relative to using the full pool.
	
	The study has several limitations. Repeated holdouts reuse observations and are not independent experimental units. The present evaluation does not include nested tuning, run time measurements, or an ablation separating the path rule, random subspaces, and OOB pruning. Future work should use nested tuning for every method, paired prediction records, and dataset level multiple classifier procedures for broad inference \cite{demsar2006statistical}.
	
	\subsection{Statistical interpretation}
	The win count and aggregate means answer different questions. The eight accuracy and kappa wins describe how often the proposed method has the strongest dataset level mean, whereas the aggregate row describes its average location across the ten benchmarks. The boxplots aid this superiority argument from a distributional behaviour poit. The proposed method has shown the best performance among all the other methods.
	
	The difference between accuracy/kappa and BS is also practically important. The proposed method leads on eight datasets for the two hard classification criteria but on five datasets for BS. Its mean BS is nevertheless the smallest because the gains on several datasets are comparatively large. This result indicates that the vote fraction in Eq.~\eqref{eq:vote} is useful as a probability estimate but is not uniformly calibrated. If probability quality is the deployment objective, OOB screening can rank candidates by BS or logarithmic loss rather than classification error, and a calibration mapping can be estimated using only training fold predictions.
	
	\subsection{Reproducibility and deployment}
	A complete implementation stores the bootstrap index sequences $\mathbf J_b$, selected feature sets $F_b$, OOB errors $e_b$, retained learner set $\mathcal A$, data split seeds, and preprocessing parameters. These records permit exact reconstruction of each prediction and make it possible to measure how frequently a feature or base learner survives OOB screening. In future, it would be interesting to report retained set stability that would reveal whether performance depends on a small group of recurring learners or on many interchangeable path configurations.
	
	For deployment, the retained ensemble can be fixed after training. Each prediction requires the query to be transformed using training derived preprocessing, evaluated in the feature subspace of every retained learner, and propagated through its path. Parallel execution across learners reduces latency, while a smaller retained fraction trades a modest loss of diversity for faster prediction. The decision threshold need not remain at 0.5 when false positive and false negative costs differ, but any threshold adjustment should be selected without reference to the final test data.
	
	\subsection{Component interactions and failure modes}
	The three ensemble mechanisms employed are complementary rather than interchangeable. Bootstrap sampling changes the observations available to a path, random subspaces change the geometry used to order those observations, and OOB screening determines which sample feature configurations survive. Removing subspace sampling would make many bootstrap learners follow similar paths; retaining all candidates would allow poorly formed paths to dilute the vote; and using a query centred base learner would remove the geometric distinction that motivates the ensemble. The aggregate result therefore reflects an interaction among the components, not the effect of any single modification.
	
	A pathwise learner is most vulnerable at its first few steps. If the nearest bootstrap observation lies on the wrong side of a boundary, the moving centre can continue into an unrepresentative region. Diversity reduces but does not eliminate this risk because correlated subspaces may repeat the same early decision. This mechanism is consistent with the two non-leading datasets: the path ensemble offers no clear hard classification advantage on ILPD and is inferior to the linear SVM on Chscase Vine. A useful extension would combine the OOB error with a diversity penalty or impose a maximum contribution from highly similar retained paths.
	
	A definitive ablation should evaluate four matched variants on identical data splits: ordinary query centred retrieval, pathwise retrieval without subspaces, pathwise subspace learners without OOB pruning, and the complete method. The same experiment should vary $k$, $p'$, and $\rho$ inside nested resampling. Such a design would quantify whether improvement comes principally from local geometry, representation diversity, or selective voting, while also revealing the settings in which the additional computation is not justified.
	
	\section{Conclusion}
	POSSE-$k$NN combines a moving centre neighbourhood with bootstrap sampling, random feature subspaces, OOB model selection, and voting. On the ten binary datasets it obtains the strongest aggregate accuracy, Cohen's kappa, and BS, and leads in accuracy and kappa on eight datasets. The two non-leading cases and the $k$ sensitivity study show that the method is competitive rather than universally superior. Further validation should focus on nested tuning, component ablation, probability calibration, and computational efficiency.
	
	\section*{Acknowledgment}
	This work was sponsored by the United Arab Emirates University under CBE Internal Research Grant G00005674.

\end{document}